\documentclass[11pt]{article}

\usepackage[final]{acl}

\usepackage{microtype}
\usepackage{inconsolata}
\usepackage{graphicx}
\usepackage{booktabs}
\usepackage{amsmath}
\usepackage{multirow}
\usepackage{xcolor}
\usepackage{latexsym}

\usepackage[english,bidi=default]{babel}

\babelfont{rm}[
  Path = ./,
  Extension = .otf,
  UprightFont = texgyretermes-regular,
  BoldFont = texgyretermes-bold,
  ItalicFont = texgyretermes-italic,
  BoldItalicFont = texgyretermes-bolditalic,
]{TeXGyreTermes}

\babelprovide[import]{arabic}
\babelfont[*arabic]{rm}[
  Path = ./,
  Extension = .ttf,
  UprightFont = Amiri-Regular,
  BoldFont = Amiri-Bold,
  ItalicFont = Amiri-Italic,
  BoldItalicFont = Amiri-BoldItalic,
]{Amiri}

\usepackage{listings}
\usepackage{tcolorbox}
\tcbuselibrary{breakable,skins}

\title{Beyond Poetry: Can Large Language Models Generate Classical Arabic Maqamat?}

\author{
AbdulRahman A. Morsy$^{1}$ \quad Aya Zirikly$^{1,2}$ \\[6pt]
$^{1}$Department of Computer Science, School of Engineering and Applied Sciences, \\
George Washington University, Washington, DC, United States \\
$^{2}$Center for Speech and Language Processing, Whiting School of Engineering, \\
Johns Hopkins University, Baltimore, MD, United States \\[6pt]
\texttt{\{abdulrahman.morsy, aya.zirikly\}@gwu.edu}
}

\begin{document}
\maketitle

\begin{abstract}
Large language models (LLMs) have shown strong performance in creative text generation, yet their ability to produce culturally grounded and stylistically constrained literary forms remains underexplored. Prior work has focused largely on modern language varieties and poetry, while classical prose traditions such as maqama (\foreignlanguage{arabic}{مقامة}) remain largely unstudied. The maqama is a classical literary genre characterized by rhymed prose (\textit{saj'}, \foreignlanguage{arabic}{سجع}), dense rhetorical ornamentation, and episodic narrative structure, making it a challenging testbed for evaluating whether LLMs can move beyond surface fluency toward deeper literary competence. In this paper, we present the first controlled evaluation study of maqama generation with LLMs, comparing five models under zero-shot, few-shot, and rule-based prompting, and evaluating outputs through both human annotation and an LLM-as-a-judge framework across dimensions such as rhetorical richness, saj' density, structural coherence, and stylistic authenticity. Our results show that prompting strategy plays a strong role in stylistic quality: few-shot prompting most consistently improves saj' density, while its effects on rhetoric and coherence vary by model, with the strongest models (GPT-4o and GPT-5.4-mini) benefiting most from rule-based prompting on these dimensions, though zero-shot prompting yields the highest aggregate scores across all five models. We further observe systematic differences between models in stylistic alignment with Arabic maqama conventions, and corroborate our findings with a second independent LLM judge, paired statistical significance testing, and non-LLM proxy measures of saj'.
\end{abstract}

\section{Introduction}

Large language models (LLMs) have demonstrated strong capabilities in natural language generation across a wide range of tasks, including dialogue, summarization, and creative writing \cite{brown2020language,touvron2023llama}. Despite these advances, their ability to generate text under strict stylistic, rhetorical, and historically grounded constraints remains an open challenge.

This challenge becomes especially evident in classical literary traditions, where success depends not only on grammatical fluency but also on adherence to genre-specific conventions. Among these traditions, the \textit{maqama} (\foreignlanguage{arabic}{مقامة}; plural \textit{maqamat}, \foreignlanguage{arabic}{مقامات}) occupies a central place in Arabic literary heritage.

The maqama is a classical prose genre that emerged in the 10th century with \textit{Badi' al-Zaman al-Hamadhani} and later reached its most canonical form with \textit{al-Ḥariri}. It is characterized by rhymed prose (\textit{saj'}, \foreignlanguage{arabic}{سجع}), dense rhetorical ornamentation, lexical ingenuity, and episodic storytelling \cite{drory2000maqama, hameen2002maqama}. Traditionally, maqamat revolve around a narrator recounting the exploits of a witty and eloquent protagonist, often involving deception, social critique, and displays of verbal mastery. The genre became one of the clearest demonstrations of linguistic virtuosity in Arabic literary history \cite{drory2000maqama, hameen2002maqama}. To illustrate these defining stylistic and rhetorical features, we include a representative excerpt from al-Hamadhani’s \textit{al-Maqama al-Ḥamdaniyya} with English translation in Appendix~\ref{app:canonical-example}.

Maqama generation presents an especially challenging testbed for modern LLMs for three reasons. First, it requires maintaining saj', a structured rhymed prose form that demands phonological and lexical control beyond ordinary prose generation. Second, it relies heavily on rhetorical devices such as parallelism, antithesis, and semantic symmetry. Third, it requires preserving narrative coherence while sustaining a consistent classical register. These properties make maqama generation a useful benchmark for evaluating whether LLMs can move beyond surface fluency toward deeper literary competence.

While recent work has made substantial progress in controllable text generation and stylistic adaptation, including narrative planning and creative writing \cite{fan2018hierarchical,rashkin2020plotmachines}, little attention has been given to classical Arabic prose. Existing Arabic NLP research has focused primarily on Modern Standard Arabic, dialectal varieties, and poetry generation \cite{antoun2020arabert,inoue2021camelbert}, leaving highly constrained prose genres largely unexplored.

In this paper, we present the first controlled study of maqama generation using modern LLMs. We evaluate multiple models under three prompting regimes: zero-shot, few-shot, and rule-based prompting, to assess their ability to reproduce the stylistic and structural properties of the genre. To do so, we introduce a structured evaluation framework combining human annotation and LLM-as-a-judge assessment across dimensions including grammar, saj', rhetoric, coherence, creativity, relevance, and stylistic authenticity.

Our study addresses four research questions:

\begin{itemize}
    \item \textbf{RQ1:} Can modern LLMs generate coherent maqama-style prose that preserves key classical literary properties?
    \item \textbf{RQ2:} How do different prompting strategies affect stylistic fidelity and rhetorical quality?
    \item \textbf{RQ3:} Which literary dimensions are most sensitive to prompt engineering, and which remain primarily model-dependent?
    \item \textbf{RQ4:} To what extent do absolute rubric-based and pairwise comparative evaluations agree when assessing highly stylized literary generation?
\end{itemize}

By answering these questions, this work contributes the first systematic evaluation study of maqama generation, provides empirical evidence on prompt sensitivity in constrained literary generation, and offers broader insights into the limits of stylistic controllability in LLMs. We release the full set of topics, prompts, and generated outputs to support future benchmarking and comparative studies on this task.


\section{Related Work}

\paragraph{Large language models and controllable generation.}
Large language models have enabled significant progress in text generation, particularly through prompting-based control mechanisms such as few-shot learning and instruction tuning \cite{brown2020language, wei2022chain}. These approaches allow models to adapt to new tasks without parameter updates and have been widely studied in both analytical and creative settings.

\paragraph{Creative and narrative generation.}
Prior work has explored controllable storytelling and narrative generation using neural models, focusing on coherence, plot structure, and long-form generation \cite{fan2018hierarchical, rashkin2020plotmachines}. However, such studies primarily target modern genres and do not address historically grounded literary traditions.

\paragraph{Arabic NLP and classical language modeling.}
Arabic NLP has evolved from early feature-based and transformer-based models such as AraBERT and CAMeLBERT \cite{antoun2020arabert, inoue2021camelbert} to modern large language models capable of open-ended generation. 
More recent Arabic-capable or Arabic-focused LLMs include models such as ALLaM~\cite{allam2024}, Jais~\cite{jais2023}, and Falcon variants~\cite{falcon2025arabic}, which incorporate Arabic data either through multilingual pretraining or targeted dataset curation.

In parallel, general-purpose instruction-tuned LLMs such as GPT-4-class models and LLaMA-family models have demonstrated strong multilingual capabilities, including in Arabic generation, despite not being explicitly optimized for classical language varieties. These models often exhibit strong fluency in Modern Standard Arabic but varying ability to capture stylistically constrained or historically grounded forms.

While some recent work explores generative capabilities in Arabic, including poetry and stylistic generation \cite{sadallah2026instruction, alghallabi2025fann}, these efforts have primarily focused on controllable poetry generation or poetic understanding benchmarks, with classical Arabic prose and rhetorical genres remaining relatively underexplored.

\paragraph{Evaluation of generated text.}
Evaluating open-ended generation remains challenging. LLM-as-a-judge frameworks have been proposed as scalable alternatives to human evaluation \cite{zheng2023judging}, and have also been adopted in recent Arabic NLP work \cite{sadallah2026instruction}. While effective in many settings, relying solely on LLM-based evaluation may introduce biases toward surface-level fluency and make results sensitive to prompt framing.

\paragraph{Cultural and stylistic gaps.}
Recent work has shown that multilingual competence does not necessarily imply cultural competence, particularly in historically situated or tradition-bound forms of language use \cite{bagheri2026culturally}. These limitations become especially pronounced in literary genres such as the maqama, where meaning depends not only on semantic adequacy but on mastery of inherited rhetorical conventions, stylistic rhythm, and cultural-literary context.

\paragraph{Gap.}
Despite progress in controllable generation and narrative modeling, no prior work has systematically studied maqama generation or evaluated LLMs on classical Arabic rhetorical prose.


\section{Data and Experimental Setup}

In the absence of a standardized dataset for classical Arabic maqamat, we construct a controlled synthetic generation suite using five large language models: GPT-4o, GPT-5.4-mini, ALLaM-7B-Instruct, Qwen3-8B, and LLaMA-3-8B-Instruct. Model selection follows prior work on Arabic constrained and literary domains, particularly recent studies on Arabic poetry such as \cite{sadallah2026instruction}, which employ a similar mixture of proprietary and open-weight multilingual and Arabic-capable models; our objective is to evaluate representative state-of-the-art proprietary and open-weight systems under a common protocol rather than to isolate the effects of model scale or training. All models are queried under identical decoding settings, with temperature set to 0.7, a value commonly used for creative text generation and consistent with prior literary generation work \cite{peeperkorn2024temperature,lewis2021syllable}, and a maximum generation length of 2400 tokens.

We define a set of 14 thematic prompts inspired by classical maqama scenarios, including travel narratives, deception, rhetorical contests, social satire, and culturally hybrid situations, themes central to the maqama tradition and discussed in \cite{drory2000maqama, hameen2002maqama}. For each configuration ($\text{model} \times \text{topic} \times \text{prompting condition}$), we generate $N=5$ independent samples, resulting in a fully factorial synthetic dataset of maqama-style outputs stored in JSONL format with full metadata for reproducibility. We release the full topic list and generated outputs as supplementary material to support reproducibility.

\section{Methods}

\subsection{Prompting Strategies}

To study the effect of in-context conditioning on stylistic generation, we apply three prompting strategies across all models and topics:

\begin{itemize}
    \item \textbf{Zero-Shot:} Minimal instruction prompting using only the target topic.
    \item \textbf{Few-Shot:} In-context learning using exemplar classical \textit{maqama} excerpts to induce stylistic imitation.
    \item \textbf{Rule-Based:} Structured instructions enforcing classical rhetorical constraints, including saj', rhetorical devices, and narrative progression.
\end{itemize}

All full prompt templates are provided in the Appendix for reproducibility.

\subsection{Automatic Evaluation}
We evaluate model outputs using an LLM-as-a-judge framework based on Claude Sonnet (claude-sonnet-4-5-20250929). Evaluation is implemented via structured tool use, enforcing constrained outputs rather than free-form judgments.

We conduct two complementary evaluation protocols. First, an \textit{absolute evaluation} setting where each generated maqama is independently scored on a 1--5 Likert scale across multiple dimensions: grammar, saj', rhetorical richness, coherence, creativity, relevance, authenticity, and overall quality. Second, a \textit{pairwise evaluation} setting where two outputs are compared directly, and the evaluator selects a winner along with per-dimension comparative scores.

To improve robustness, evaluations are performed with temperature set to 0, and included randomized pairing order to mitigate positional bias. In the pairwise setting, the identity of the candidates is randomly swapped and later remapped to their original model identities.

The full evaluation prompts used for both absolute and pairwise settings are provided in the Appendix.

\subsection{Human Evaluation}
To complement automatic evaluation, we conduct a human evaluation study with two annotators who are native Arabic speakers with a strong background in classical Arabic language and literary studies. We construct a balanced set of 60 maqamat (30 pairwise comparisons) sampled to cover all ten model pairs across all three prompting strategies.

The evaluation includes two complementary comparison settings. First, annotators perform cross-model comparisons, where outputs from different models are directly compared under identical prompts. Second, to isolate the effect of prompting strategy, annotators additionally compare outputs generated by the same model across different prompting conditions (zero-shot, few-shot, and rule-based).

Annotators are asked to select a preferred output in each pair and provide a brief justification focusing on overall literary quality. To mitigate position and cognitive bias, text pairs are presented in a fully randomized, double-blind layout with all model names and prompt metadata stripped. Annotators perform a forced-choice pairwise preference selection based on structural adherence to classical maqama style (focusing heavily on saj' precision, rhetorical flair, and narrative cohesion). Inter-annotator agreement is quantified using Cohen's Kappa ($\kappa$).

\begin{table*}[t]
\centering
\small
\caption{Average Claude-Sonnet-4.5-as-judge evaluation scores (1--5) for GPT-4o and GPT-5.4-mini across prompting conditions and quality dimensions.}
\label{tab:scores}
\begin{tabular}{llccccccc}
\toprule
\textbf{Model} & \textbf{Condition} & \textbf{Coherence} & \textbf{Creativity} & \textbf{Grammar} & \textbf{Overall} & \textbf{Relevance} & \textbf{Rhetoric} & \textbf{Saj'} \\
\midrule
\multirow{3}{*}{GPT-4o}
 & Few-shot   & 3.95 & \textbf{4.02} & 3.98 & 4.05 & 4.52 & \textbf{4.07} & \textbf{3.74} \\
 & Rules      & 3.83 & 3.83 & 3.93 & 4.07 & 4.57 & 3.83 & 3.69 \\
 & Zero-shot  & \textbf{4.31} & 3.90 & \textbf{4.33} & \textbf{4.19} & \textbf{4.74} & \textbf{4.07} & 3.57 \\
\midrule
\multirow{3}{*}{GPT-5.4-mini}
 & Few-shot   & 3.90 & 3.90 & 3.86 & 3.74 & 4.43 & 3.90 & 3.57 \\
 & Rules      & 3.62 & \textbf{3.98} & 3.71 & 3.88 & 4.50 & \textbf{4.17} & \textbf{4.14} \\
 & Zero-shot  & \textbf{4.12} & 3.69 & \textbf{4.26} & \textbf{3.98} & \textbf{4.62} & 3.93 & 3.55 \\
\bottomrule
\end{tabular}
\end{table*}

\section{Results}
\label{sec:results}

We evaluate model outputs using both absolute rubric-based scoring and pairwise comparative judgment, enabling both independent quality assessment and relative ranking. Full supplementary analyses are reported in Appendix~\ref{app:additional-results}.

\subsection{Absolute Evaluation}

Table~\ref{tab:absolute-main} reports the mean absolute scores across all evaluation dimensions.

\begin{table*}[t]
\centering
\small
\caption{Mean absolute evaluation score per dimension for each model, averaged across all three prompting conditions.}
\label{tab:absolute-main}
\begin{tabular}{lcccccccc}
\toprule
Model & Grammar & Saj' & Rhetoric & Coherence & Creativity & Relevance & Authenticity & Overall \\
\midrule
GPT-5.4-mini & 5.00 & 4.44 & 4.17 & 4.82 & 3.88 & 4.95 & 4.14 & \textbf{4.17} \\
GPT-4o & 4.06 & 2.59 & 2.76 & 3.47 & 2.15 & 3.55 & 2.21 & \textbf{2.65} \\
ALLaM & 3.23 & 1.53 & 1.68 & 1.93 & 1.32 & 2.13 & 1.26 & \textbf{1.57} \\
Qwen & 1.76 & 1.11 & 1.30 & 1.28 & 1.31 & 1.69 & 1.04 & \textbf{1.19} \\
LLaMA & 1.77 & 1.26 & 1.05 & 1.07 & 1.11 & 1.47 & 1.02 & \textbf{1.05} \\
\bottomrule
\end{tabular}
\end{table*}

GPT-5.4-mini substantially outperforms all baselines, particularly in maqama-specific dimensions such as saj', rhetoric, and authenticity, where weaker models collapse despite retaining moderate grammaticality. This suggests that literary imitation requires capabilities beyond surface-level fluency.

Notably, GPT-4o achieves relatively strong grammatical and coherence scores but remains significantly weaker in authenticity, indicating partial stylistic imitation without full genre realization. Despite being Arabic-specialized, ALLaM underperforms in most literary dimensions, suggesting that general Arabic fluency alone does not translate to classical maqama competence.

Beyond mean scores, GPT-5.4-mini also achieves the lowest variance among the two strong-performing models (Overall std.\ 0.38 vs.\ 0.53 for GPT-4o), indicating not only higher but more consistent quality across topics and samples; open-weight models exhibit both lower performance and reduced stability. A full stability breakdown for all five models is provided in Appendix~\ref{app:stability}.

\subsection{Prompting Effects}

To assess prompt sensitivity, we compare model performance across three prompting conditions.

\begin{table}[t]
\centering
\small
\caption{Mean overall score by prompting condition.}
\label{tab:conditions}
\setlength{\tabcolsep}{2.5pt}
\begin{tabular}{lccccc}
\toprule
Condition & ALLaM & GPT-4o & GPT-5.4-mini & LLaMA & Qwen \\
\midrule
Zero-shot & 1.44 & 2.33 & 4.16 & 1.00 & 1.51 \\
Few-shot  & 1.60 & \textbf{2.87} & 4.11 & \textbf{1.14} & 1.01 \\
Rules     & 1.66 & 2.74 & \textbf{4.24} & 1.01 & 1.04 \\
\bottomrule
\end{tabular}
\end{table}

Few-shot prompting yields the largest gains for GPT-4o (+0.54), while GPT-5.4-mini remains comparatively stable across all conditions, suggesting lower dependence on prompt scaffolding. This pattern indicates that stronger models internalize stylistic constraints more effectively, whereas weaker models benefit more from explicit examples.

Table~\ref{tab:scores} reports the full per-dimension breakdown for these two strongest models across prompting conditions. Both GPT-4o and GPT-5.4-mini achieve their best rhetoric and coherence scores under rule-based or few-shot prompting, whereas the aggregate result across all five models (Table~\ref{tab:dimension-sensitivity}) favors zero-shot prompting on these same dimensions, since the weaker open-weight models are destabilized by structured or few-shot prompting. Table~\ref{tab:model-prompt-dynamics} (Appendix~\ref{app:model-prompt-dynamics}) reports this fine-grained breakdown for all five models and complements, rather than subsumes, the aggregate Overall score reported in Table~\ref{tab:conditions}.

A full prompt sensitivity breakdown appears in Appendix~\ref{app:prompt-sensitivity}.

\subsection{Pairwise Comparison}

We complement absolute scoring with pairwise comparative evaluation.

\begin{table*}[t]
\centering
\small
\caption{Pairwise win-rate matrix.}
\label{tab:pairwise}
\begin{tabular}{lccccc}
\toprule
Model & ALLaM & GPT-4o & GPT-5.4-mini & LLaMA & Qwen \\
\midrule
ALLaM & -- & 0.00 & 0.00 & \textbf{0.85} & \textbf{0.77} \\
GPT-4o & \textbf{1.00} & -- & 0.00 & \textbf{1.00} & \textbf{1.00} \\
GPT-5.4-mini & \textbf{1.00} & \textbf{1.00} & -- & \textbf{1.00} & \textbf{1.00} \\
LLaMA & 0.15 & 0.00 & 0.00 & -- & \textbf{0.53} \\
Qwen & 0.23 & 0.00 & 0.00 & 0.48 & -- \\
\bottomrule
\end{tabular}
\end{table*}

As shown in Table~\ref{tab:pairwise}, pairwise judgments produce a clean hierarchy. GPT-5.4-mini achieves a perfect win rate against all competing systems, while GPT-4o consistently defeats all non-GPT baselines. Among open models, ALLaM substantially outperforms both Qwen and LLaMA, supporting the hypothesis that Arabic specialization provides some advantage, even when insufficient for strong maqama generation. One pair is not statistically distinguishable from chance: LLaMA's win rate over Qwen is 52.4\% (bootstrap 95\% CI [0.381, 0.667]), so the relative order of the two lowest-ranked models should be interpreted with caution despite their separation in Elo.

\subsection{Global Ranking}

To derive a global comparative ranking, we compute Elo scores over pairwise outcomes.

\begin{table}[t]
\centering
\small
\caption{Elo ranking derived from pairwise comparisons.}
\label{tab:elo}
\begin{tabular}{lcc}
\toprule
Rank & Model & Elo \\
\midrule
1 & GPT-5.4-mini & \textbf{1980.40} \\
2 & GPT-4o & \textbf{1638.93} \\
3 & ALLaM & \textbf{1261.28} \\
4 & Qwen & \textbf{1099.37} \\
5 & LLaMA & \textbf{1020.02} \\
\bottomrule
\end{tabular}
\end{table}

As shown in Table~\ref{tab:elo}, the Elo results reveal four distinct performance tiers, with large separations between adjacent systems. The strongest gap appears between GPT-5.4-mini and GPT-4o (+341 Elo), highlighting a substantial qualitative leap in literary competence.

To assess the robustness of this ranking, we conducted bootstrap confidence interval estimation and paired Wilcoxon signed-rank significance testing on all 1,050 absolute judgments, matched by condition, topic, and sample. Every pairwise difference in mean Overall score between the five models is significant at $p<0.001$, surviving Holm-Bonferroni correction across all ten pairwise comparisons, indicating that the reported ranking is statistically robust with the single exception noted above (LLaMA vs.\ Qwen). We further re-evaluated the full corpus (1,050 absolute judgments and 420 pairwise comparisons) using Gemini 3.5 Flash as an independent second LLM judge, following the practice of using an external judge model to avoid bias toward any evaluated model family \cite{sadallah2026instruction}. Agreement between the two judges is strong: Spearman $\rho=0.878$ on absolute Overall scores (per-dimension $\rho=0.746$--$0.945$), 93.3\% raw agreement and Cohen's $\kappa=0.907$ on pairwise winners, and $\rho=0.90$ agreement on the resulting five-model ranking, differing by only a single adjacent swap between the third- and fourth-ranked models (ALLaM and Qwen). Combined, these results indicate that the reported model ranking is stable across statistical testing and judge model, not solely dependent on Claude Sonnet 4.5.

Agreement between absolute and pairwise rankings is analyzed in Appendix~\ref{app:agreement}.

\subsection{Human Evaluation}
Human evaluation shows strong consistency between annotators, with a raw agreement of 0.9667 and a Cohen’s kappa of 0.9231, indicating near-perfect inter-annotator reliability. This high agreement suggests that preferences among maqama outputs are highly stable and that stylistic quality is clearly discernible to expert Arabic readers.

Across evaluations, both annotators consistently preferred few-shot and rule-based prompting outputs over zero-shot generations, highlighting the importance of explicit stylistic conditioning for achieving classical maqama fidelity. Annotators emphasized that improvements were primarily driven by better control of saj' structure, rhetorical density, and narrative cohesion.

A notable qualitative finding was the presence of hallucinated or structurally invalid content in several open-source model outputs, including repetitive passages and occasional fabricated religious quotations attributed to the Qur’an and Hadith.

Overall, closed-source models significantly outperformed open-source alternatives in perceived maqama quality. GPT-5.4-mini achieved a consistent win across all pairwise comparisons, followed by GPT-4o, which also showed strong and stable performance against all remaining models. This ranking closely aligns with the LLM-as-a-judge results using Claude Sonnet 4.5, indicating strong agreement between automated and human evaluation signals.

\subsection{Non-LLM Validation of Saj'}
\label{sec:saj-proxy}

To support the literary evaluation with evidence independent of LLM judgment, we implemented two deterministic, non-LLM proxy measures for saj': rhyme-ending consistency (the proportion of adjacent clause pairs sharing the same final letter) and cadence regularity (based on the coefficient of variation in clause length), computed directly on all 1,050 generated maqamat. Rhyme-ending consistency correlates significantly with the LLM judge's saj' scores (Spearman's $\rho=0.328$, $p<0.001$), indicating that the judge's assessments are at least partly grounded in measurable stylistic properties rather than arbitrary. These proxies are intentionally lightweight and are not intended to replace literary evaluation; a fuller computational assessment of saj' would require substantially richer linguistic modeling, which we leave to future work.

\subsection{Diacritization}
\label{sec:diacritization}

Motivated by the observation that classical maqamat are conventionally diacritized, we additionally analyzed the density of diacritic marks in the generated corpus. Few-shot prompting consistently produced substantially more diacritized output than zero-shot or rule-based prompting across all evaluated models (for example, Qwen increased from 30.6 to 77.3 diacritic marks per 100 Arabic letters), suggesting that models largely acquire this behavior by imitating the diacritized in-context exemplars rather than through an intrinsic tendency to vocalize the text. Evaluating the correctness of generated diacritics, as opposed to their mere presence, requires gold-standard diacritized references and dedicated evaluation metrics; we leave diacritization-aware prompting and rigorous evaluation of vocalization quality, including its interaction with literary devices such as \textit{jinās}, to future work.

\section{Discussion}
\label{sec:discussion}

Our findings reveal that maqama generation remains a highly demanding literary task for current language models. While strong general language ability is necessary, it is not sufficient for successful maqama generation, which requires mastery of genre-specific constraints such as saj', rhetorical ornamentation, and stylistic authenticity.

GPT-5.4-mini demonstrates the strongest and most consistent performance among all the evaluated models, suggesting robust internalization of maqama-style constraints. By contrast, GPT-4o produces fluent and coherent Arabic but remains weaker in authenticity and saj', indicating that surface fluency does not guarantee successful classical style imitation. Open-weight models perform substantially worse overall, showing that general Arabic capability alone remains insufficient for this genre.

We note that differences may partially reflect instruction-following fidelity and pretraining exposure to classical Arabic registers, rather than purely generative capacity in isolation.

Prompting affects models unevenly. Most models benefit from structured prompting, especially few-shot examples, but these gains are concentrated in surface-level features. Saj' emerges as the most prompt-sensitive dimension, indicating that rhythmic and rhymed prose can be partially induced through prompting. In contrast, creativity and rhetoric remain relatively stable, suggesting that deeper literary competence is largely model-bound.

Interestingly, ALLaM exhibits non-monotonic behavior, where prompting sometimes degrades performance rather than improving it. This may indicate interference between imposed stylistic constraints and the model’s existing Arabic generation priors.

Finally, the strong agreement between absolute and pairwise evaluation suggests that the observed ranking is robust across evaluation paradigms, reinforcing confidence in the overall findings.

\section{Conclusion}
\label{sec:conclusion}

In this paper, we evaluated the ability of modern language models to generate maqama-style Arabic text using a dual evaluation framework combining absolute rubric-based scoring and pairwise comparison.

Our results establish a clear performance hierarchy, with GPT-5.4-mini achieving the strongest results, followed by GPT-4o, while open-weight models remain significantly behind. We further show that prompting can improve weaker models, particularly in stylistic dimensions such as saj', but has limited impact on stronger models.

These findings suggest that maqama generation depends not only on grammatical fluency, but also on deeper stylistic and rhetorical competence that remains difficult for most models to acquire.

Future work can extend this setting toward longer-form maqamat with richer narrative structure and stronger alignment to classical exemplars such as al-Hamadhani and al-Hariri, enabling evaluation of both local stylistic fidelity and long-range literary coherence.

\section{Limitations}
\label{sec:limitations}

This study has several limitations that should be considered when interpreting the results.

\paragraph{Scope of evaluation.}
We evaluate short-form maqama-style generations over 14 curated topics. This setup does not capture the full structural and thematic complexity of classical maqama literature, which involves longer narratives and richer discourse development.

\paragraph{Evaluation method.}
We rely mainly on LLM-as-a-judge with a structured rubric, supported by pairwise comparisons and limited human evaluation. While rankings are consistent across methods, automated judgments may still reflect model-specific bias, and human evaluation is not large enough to serve as a full reference standard.

\paragraph{Pairwise sampling.}
Pairwise results are computed over a stratified sample of 420 comparisons rather than exhaustively over all pairs, introducing potential sampling variance.

\section*{Ethics Statement}

This work involves the generation of culturally and historically grounded literary text. While we do not observe direct harmful outputs, we note that models may reproduce simplified or stereotypical representations of cultural identities. Care must be taken when interpreting generated narratives as they reflect statistical learning rather than historical understanding.

\section*{AI Usage Statement}

We acknowledge the use of large language models (specifically ChatGPT) to assist in manuscript preparation through writing support, language refinement, and verification of parts of the reported analyses. All generated content was carefully reviewed and validated by the authors. The research hypotheses, experimental design, implementation, and final scientific conclusions were conceived and conducted entirely by the authors.

\bibliography{custom}

\appendix
\section{Canonical Classical Maqama Example}
\label{app:canonical-example}

To illustrate the stylistic target of our generation setup, we provide an excerpt from al-Hamadhani’s \textit{al-Maqama al-Ḥamdaniyya} (\foreignlanguage{arabic}{المقامة الحمدانية}), one of the earliest and most influential examples of the genre. This excerpt demonstrates several defining maqama properties: framed narration, rhymed prose (saj'), dense rhetorical ornamentation, lexical symmetry, and performative eloquence.

\paragraph{Arabic excerpt.}

\begin{quote}
\foreignlanguage{arabic}{
حَدَّثَنَا عِيسَى بْنُ هِشَامٍ قَالَ: حَضَرْنَا مَجْلِسَ سَيْفِ الدَّوْلَةِ بْنِ حَمْدَانَ يَوْمًا، وَقَدْ عُرِضَ عَلَيْهِ فَرَسٌ مَتَى مَا تَرَقَّتِ العَيْنُ فِيهِ تَسَهَّلَ، فَلَحَظَتْهُ الجَمَاعَةُ، وَقَالَ سَيْفُ الدَّوْلَةِ: أَيُّكُمْ أَحْسَنَ صِفَتَهُ، جَعَلْتُهُ صِلَتَهُ.

فَكُلٌّ جَهَدَ جَهْدَهُ، وَبَذَلَ مَا عِنْدَهُ، فَقَالَ أَحَدُ خَدَمِهِ: أَصْلَحَ اللهُ الأَمِيرَ! رَأَيْتُ بِالأَمْسِ رَجُلًا يَطَأُ الفَصَاحَةَ بِنَعْلَيْهِ، وَتَقِفُ الأَبْصَارُ عَلَيْهِ، يَسْأَلُ النَّاسَ، وَيَسْقِي اليَأْسَ.

فَأُحْضِرَ، وَقَدْ عَلَيْهِ طِمْرَانِ قَدْ أَكَلَ الدَّهْرُ عَلَيْهِمَا وَشَرِبَ، فَقَالَ سَيْفُ الدَّوْلَةِ: بَلَغَتْنَا عَنْكَ عَارِضَةٌ، فَاعْرِضْهَا فِي هَذَا الفَرَسِ وَوَصْفِهِ.
}
\end{quote}

\paragraph{English translation (excerpt).}

\begin{quote}
Isa ibn Hisham narrated to us, saying: One day we attended the court of Sayf al-Dawla ibn Hamdan, when a horse was presented before him, so splendid that any eye rising to behold it found delight. The gathering gazed upon it, and Sayf al-Dawla said: “Whichever of you describes it best, I shall grant it to him.”

Each man exerted himself and offered what he knew. Then one of his servants said: “May God preserve the prince! Yesterday I saw a man who tramples eloquence beneath his sandals, toward whom all eyes turn, who questions men and pours despair upon them.”

So he was brought in at once, clothed in two worn garments upon which time had eaten and drunk. Then Sayf al-Dawla said to him: “We have heard of your eloquence; now display it in describing this horse.”
\end{quote}

This excerpt illustrates the central mechanics of the maqama genre: the framed narrative voice, the staged performance of eloquence, and the dense use of saj' and rhetorical embellishment. These features motivate the dimensions used in our evaluation framework, particularly authenticity, rhetoric, coherence, and saj' density.

\section{Generation Prompt Templates}
\label{app:generation-prompts}

This appendix provides the exact prompting templates used for synthetic maqama generation. Prompts are shown in both Arabic (original) and English (translation) for reproducibility.

\subsection{Zero-Shot Prompt}

\paragraph{Arabic}
\begin{quote}
\foreignlanguage{arabic}{
اكتب مقامة عربية فصيحة حول الموضوع التالي:

[TOPIC]

اكتب النص مباشرة بأسلوب أدبي عربي.

مهم جدًّا: ابدأ الكتابة مباشرة باللغة العربية، ولا تكتب أي مقدمات أو ملاحظات توضيحية باللغة الإنجليزية.
}
\end{quote}

\paragraph{English Translation}
\begin{quote}\ttfamily
Write an eloquent Arabic maqama on the following topic:

[TOPIC]

Write the text directly in an Arabic literary style.

Very important: begin writing directly in Arabic, and do not include any introductions or explanatory notes in English.
\end{quote}

\subsection{Few-Shot Prompt}

\paragraph{Arabic}
\begin{quote}
\foreignlanguage{arabic}{
أنت كاتب مقامات عربي بارع.

فيما يلي أمثلة على أسلوب المقامات:

\{EXAMPLES\}

الآن اكتب مقامة جديدة مستوحاة من هذا الأسلوب حول الموضوع التالي:

[TOPIC]

يجب الحفاظ على الفصاحة والبلاغة ووحدة الأسلوب.
}
\end{quote}

\paragraph{English Translation}
\begin{quote}\ttfamily
You are a skilled Arabic maqama writer.

Below are examples of maqama style:

\{EXAMPLES\}

Now write a new maqama inspired by this style on the following topic:

[TOPIC]

You must preserve eloquence, rhetorical richness, and stylistic consistency.
\end{quote}

\subsection{Rule-Based Prompt}

\paragraph{Arabic}
\begin{quote}
\foreignlanguage{arabic}{
أنت كاتب مقامات عربي فصيح على مستوى كبار أدباء السجع والبلاغة الكلاسيكية.

مهمتك كتابة مقامة عربية بليغة حول الموضوع التالي:

[TOPIC]

قواعد إلزامية (MUST FOLLOW)

1) الأسلوب: مقامة مكتوبة بأسلوب عربي كلاسيكي فصيح يعتمد على السجع المنتظم.

2) السجع: استخدمه بشكل متكرر وطبيعي عبر قوافٍ نثرية واضحة.

3) البلاغة: تضمين الجناس، الطباق، المقابلة، والتشبيه الأدبي التراثي.

4) البنية: تتضمن مقدمة تمهيدية، عقدة أو حدثًا محوريًّا، وخاتمة ذات حكمة.

5) الطول: بين 600 و800 كلمة تقريبًا.

ممنوعات

- ممنوع السرد البسيط الخالي من البلاغة.
- ممنوع العامية.
- ممنوع التفسير خارج النص.
- ممنوع العناوين الشارحة.

OUTPUT

اكتب المقامة مباشرة دون أي مقدمة تفسيرية.
}
\end{quote}

\paragraph{English Translation}
\begin{quote}\ttfamily
You are an eloquent Arabic maqama writer at the level of the great masters of rhymed prose and classical rhetoric.

Your task is to write an eloquent Arabic maqama on the following topic:

[TOPIC]

Mandatory Rules (MUST FOLLOW)

1) Style: The maqama must be written in an eloquent classical Arabic style relying on regular rhymed prose (saj').

2) Saj': Use it frequently and naturally through clear prose rhyme.

3) Rhetoric: Include paronomasia, antithesis, parallelism, and classical similes.

4) Structure: Include an opening introduction, a central conflict or event, and a closing moral reflection.

5) Length: Approximately 600--800 words.

Forbidden

- Simple plain narration lacking rhetoric.
- Colloquial Arabic.
- Any explanatory text outside the narrative.
- Explicit section headings.

OUTPUT

Write the maqama directly without any explanatory introduction.
\end{quote}

\section{Evaluation Protocol}
\label{app:evaluation-prompts}

We use a two-stage LLM-as-a-judge pipeline implemented with Claude Sonnet 4.5 at temperature zero.

\subsection{Absolute Evaluation}

\paragraph{Arabic}

\begin{quote}
\foreignlanguage{arabic}{
أنت ناقدٌ أدبيٌّ متمرّس في التراث العربي، تنظر في النصوص بعين البلاغيين الأوائل، وتزنها بميزان الفصاحة والبيان كما فعل الجرجاني وابن الأثير.

بين يديك نصٌّ كُتب محاكاةً لفنّ المقامة العربية على نهج بديع الزمان الهمذاني والحريري، وهو فنّ يقوم على السجع، والحكاية، والمفارقة، وجزالة اللفظ، وحسن السبك.

اقرأ النص بوصفه محاولةً للانتماء إلى هذا التراث البلاغي، ولا تحاكمه بمعايير الكتابة الحديثة، بل بميزان المقامة الكلاسيكية وأصولها.

قيّم النص وفق المعايير الآتية:

• جودة السجع: أهو مطبوع منسجم أم متكلّف متعسّف؟

• روح المقامة: هل يحفظ طابع الحكاية والمشهد والتحوّل؟

• اللغة: أهي جزلة تراثية أم يغلب عليها طابع العصر؟

• الإيقاع: هل للنص موسيقى داخلية وانسجام لفظي؟

• البلاغة: هل فيه صور بيانية حيّة أم زخرف لفظي مجرد؟

• التماسك: هل يجري النص في نسق سردي متصل أم يبدو مفككًا؟

واعتمد سلّم التقدير الآتي من 1 إلى 5:

1 = بعيدٌ جدًّا عن روح المقامة

2 = محاولة ضعيفة لا تبلغ مستوى المحاكاة

3 = نص متوسط تظهر فيه بعض سمات المقامة

4 = نص قوي يقترب من الأسلوب التراثي

5 = نص بالغ الجودة يكاد يُظنّ من صميم التراث

الموضوع:

\{topic\}

النص:

\{text\}

أصدر حكمك النقدي، ثم سجّل درجاتك وفق المعايير المذكورة.
}
\end{quote}

\paragraph{English Translation}

\begin{quote}\ttfamily
You are a literary critic deeply versed in the Arabic tradition, examining texts through the lens of the early masters of rhetoric and weighing them by the standards of eloquence and expression, as did al-Jurjani and Ibn al-Athir.

Before you is a text written in imitation of the Arabic maqama tradition in the style of al-Hamadhani and al-Hariri, a genre built upon rhymed prose, narrative movement, paradox, lexical grandeur, and stylistic elegance.

Read this text as an attempt to belong to that rhetorical tradition. Do not judge it by the standards of modern writing, but by the conventions and literary principles of the classical maqama.

Evaluate the text according to the following criteria:

• Quality of saj': Is it natural and harmonious, or forced and artificial?

• Spirit of the maqama: Does it preserve narrative movement, scene-building, and transformation?

• Language: Is it elevated and classical, or marked by modern linguistic tendencies?

• Rhythm: Does the text sustain internal musicality and verbal harmony?

• Rhetoric: Does it contain vivid figurative imagery, or merely superficial ornamentation?

• Coherence: Does the narrative flow cohesively, or does it feel fragmented?

Use the following rating scale from 1 to 5:

1 = Very far from the spirit of the maqama

2 = A weak attempt that fails as imitation

3 = A moderate text showing some maqama features

4 = A strong text approaching the classical style

5 = An exceptional text that could almost be mistaken for authentic classical prose

Topic:

\{topic\}

Text:

\{text\}

Deliver your literary judgment, then record your scores according to the criteria above.
\end{quote}

\section{Additional Evaluation Analyses}
\label{app:additional-results}

This section provides extended analyses of model behavior across stability, prompting strategies, pairwise judgments, and fine-grained literary dimensions. These results complement the main findings in Section~\ref{sec:results}.

\subsection{Stability Analysis}
\label{app:stability}

We first examine the stability of model outputs across all evaluation instances by measuring the mean and standard deviation of overall scores. This provides an indication of both average performance and consistency. The results are shown in Table~\ref{tab:stability}.

\begin{table}[h]
\centering
\small
\caption{Mean and standard deviation of overall scores.}
\label{tab:stability}
\begin{tabular}{lcc}
\toprule
Model & Mean & Std \\
\midrule
GPT-5.4-mini & 4.17 & 0.38 \\
GPT-4o & 2.65 & 0.53 \\
ALLaM & 1.57 & 0.50 \\
Qwen & 1.19 & 0.39 \\
LLaMA & 1.05 & 0.22 \\
\bottomrule
\end{tabular}
\end{table}

As shown in Table~\ref{tab:stability}, GPT-5.4-mini achieves both the highest mean and relatively low variance among strong-performing models, indicating consistent high-quality maqama generation. GPT-4o shows moderate variance, while open-weight models exhibit both lower performance and reduced stability.

\subsection{Prompt Sensitivity}
\label{app:prompt-sensitivity}

We next analyze the sensitivity of models to prompting strategies (zero-shot, rule-based, and few-shot). Table~\ref{tab:prompt-sensitivity} reports the best and worst-performing conditions per model.

\begin{table}[h]
\centering
\small
\caption{Best and worst prompting conditions.}
\label{tab:prompt-sensitivity}
\begin{tabular}{lccc}
\toprule
Model & Best & Worst & $\Delta$ \\
\midrule
ALLaM & Rules & Zero-shot & 0.21 \\
GPT-4o & Few-shot & Zero-shot & 0.54 \\
GPT-5.4-mini & Rules & Few-shot & 0.13 \\
LLaMA & Few-shot & Zero-shot & 0.14 \\
Qwen & Zero-shot & Few-shot & 0.50 \\
\bottomrule
\end{tabular}
\end{table}

As reported in Table~\ref{tab:prompt-sensitivity}, prompt sensitivity varies significantly across models. GPT-4o and Qwen benefit substantially from structured prompting, indicating partial reliance on external scaffolding. In contrast, GPT-5.4-mini remains highly stable, suggesting strong internalization of maqama-style constraints. Interestingly, ALLaM shows only marginal gains from prompting and even slight degradation under few-shot settings, suggesting possible over-constraining effects. Detailed analyses of evaluated dimensions is provided in~\ref{app:model-prompt-dynamics}.

\subsection{Agreement Between Evaluation Protocols}
\label{app:agreement}

We measure agreement between absolute scoring and pairwise ranking using Spearman’s rank correlation coefficient. The results are shown in Table~\ref{tab:agreement}.

\begin{table}[h]
\centering
\small
\caption{Correlation between absolute and pairwise rankings.}
\label{tab:agreement}
\begin{tabular}{lc}
\toprule
Metric & Value \\
\midrule
Spearman's $\rho$ & \textbf{1.00} \\
$p$-value & \textbf{$<$ 0.001} \\
\bottomrule
\end{tabular}
\end{table}

As shown in Table~\ref{tab:agreement}, the perfect monotonic correlation indicates that both evaluation paradigms yield fully consistent rankings, reinforcing the robustness of the evaluation framework.

\subsection{Dimension Difficulty}
\label{app:dimension-difficulty}

We next analyze the intrinsic difficulty of each literary dimension by averaging scores across all models and conditions. The results are reported in Table~\ref{tab:dimension-difficulty}, which reveals a clear hierarchy of difficulty: grammatical correctness is relatively well handled across models, whereas higher-level literary attributes such as authenticity and creativity remain significantly underdeveloped. This suggests that maqama generation is constrained more by stylistic and narrative competence than by surface-level linguistic proficiency.

\begin{table}[h]
\centering
\small
\caption{Average score by dimension.}
\label{tab:dimension-difficulty}
\begin{tabular}{lc}
\toprule
Dimension & Mean \\
\midrule
Authenticity & 1.93 \\
Creativity & 1.95 \\
Overall & 2.13 \\
Saj' & 2.19 \\
Rhetoric & 2.19 \\
Coherence & 2.51 \\
Relevance & 2.76 \\
Grammar & 3.16 \\
\bottomrule
\end{tabular}
\end{table}

\subsection{Model-wise Prompt Dynamics}
\label{app:model-prompt-dynamics}

We now turn to a fine-grained analysis of how prompting strategies affect each model across individual literary dimensions. Table~\ref{tab:model-prompt-dynamics} presents results for zero-shot, few-shot, and rule-based prompting across all models and dimensions.

\begin{table*}[h]
\centering
\small
\caption{Fine-grained evaluation across models, prompting strategies, and literary dimensions.}
\label{tab:model-prompt-dynamics}
\begin{tabular}{llccccccc}
\toprule
Model & Condition & Grammar & Saj' & Rhetoric & Coherence & Creativity & Relevance & Authenticity \\
\midrule
ALLaM & Zero-shot & 3.61 & 1.23 & 1.67 & 2.34 & 1.24 & 2.13 & 1.09 \\
      & Few-shot  & 2.96 & 1.70 & 1.64 & 1.69 & 1.39 & 2.14 & 1.44 \\
      & Rules     & 3.13 & 1.66 & 1.71 & 1.76 & 1.34 & 2.11 & 1.24 \\
\midrule
GPT-4o & Zero-shot & 4.00 & 2.24 & 2.53 & 3.29 & 2.10 & 3.34 & 2.04 \\
       & Few-shot  & 4.11 & 2.87 & 2.91 & 3.54 & 2.19 & 3.74 & 2.44 \\
       & Rules     & 4.06 & 2.66 & 2.83 & 3.59 & 2.16 & 3.56 & 2.14 \\
\midrule
GPT-5.4-mini & Zero-shot & 5.00 & 4.29 & 4.16 & 4.79 & 3.79 & 4.94 & 4.10 \\
             & Few-shot  & 5.00 & 4.61 & 4.11 & 4.83 & 3.89 & 4.96 & 4.10 \\
             & Rules     & 5.00 & 4.43 & 4.24 & 4.84 & 3.97 & 4.96 & 4.21 \\
\midrule
LLaMA & Zero-shot & 1.67 & 1.00 & 1.00 & 1.14 & 1.06 & 1.56 & 1.00 \\
      & Few-shot  & 1.94 & 1.77 & 1.14 & 1.03 & 1.24 & 1.50 & 1.06 \\
      & Rules     & 1.69 & 1.01 & 1.01 & 1.04 & 1.01 & 1.34 & 1.00 \\
\midrule
Qwen & Zero-shot & 2.26 & 1.16 & 1.73 & 1.80 & 1.71 & 2.19 & 1.10 \\
     & Few-shot  & 1.60 & 1.10 & 1.06 & 1.00 & 1.14 & 1.53 & 1.00 \\
     & Rules     & 1.41 & 1.06 & 1.10 & 1.04 & 1.09 & 1.34 & 1.01 \\
\bottomrule
\end{tabular}
\end{table*}

As shown in Table~\ref{tab:model-prompt-dynamics}, GPT-5.4-mini is largely invariant across prompting conditions, suggesting strong internalization of maqama-style constraints, yet, it still shows improvement in saj' and creativity with few-shot and rule prompting. GPT-4o exhibits moderate sensitivity, particularly in stylistic dimensions such as saj' and rhetoric, where few-shot prompting consistently improves performance. In contrast, ALLaM shows non-monotonic behavior, with occasional degradation under few-shot prompting, suggesting possible over-constraining effects. Open-weight models (LLaMA and Qwen) remain unstable and show limited consistent gains from prompting, indicating insufficient capacity for reliable stylistic control.

\subsection{Dimension-level Prompt Sensitivity}
\label{app:dimension-sensitivity}

Finally, we aggregate across models to analyze which literary dimensions are most sensitive to prompting strategies. The results are shown in Table~\ref{tab:dimension-sensitivity}.

\begin{table}[h]
\centering
\small
\caption{Sensitivity of literary dimensions to prompting strategies (higher $\Delta$ indicates stronger prompt dependence).}
\label{tab:dimension-sensitivity}
\begin{tabular}{lccc}
\toprule
Dimension & Best Condition & Worst Condition & $\Delta$ \\
\midrule
Saj' & Few-shot & Zero-shot & 0.429 \\
Coherence & Zero-shot & Few-shot & 0.254 \\
Grammar & Zero-shot & Rules & 0.251 \\
Relevance & Zero-shot & Rules & 0.169 \\
Authenticity & Few-shot & Zero-shot & 0.143 \\
Creativity & Zero-shot & Rules & 0.066 \\
Overall & Few-shot & Zero-shot & 0.060 \\
Rhetoric & Zero-shot & Few-shot & 0.043 \\
\bottomrule
\end{tabular}
\end{table}

As shown in Table~\ref{tab:dimension-sensitivity}, Saj' is the most prompt-sensitive dimension, indicating that rhymed prose structure can be effectively influenced through examples. Grammar and coherence show moderate sensitivity, suggesting partial controllability of surface linguistic structure. In contrast, rhetoric and creativity exhibit minimal variation, implying that deeper literary competence is largely model-dependent rather than prompt-driven.

\end{document}